\documentclass{article}
\pdfoutput=1 
\usepackage{subfigure,typearea}
\usepackage{caption}
\usepackage{graphicx}
\usepackage{algpseudocode}
\usepackage{algorithm}
\usepackage{hyperref}
\usepackage{amsmath,amssymb}
\areaset{14.5cm}{23cm}

\title{Supervised learning of short and high-dimensional temporal sequences for life science measurements}
\author{F.-M. Schleif\,$^{1}$ \and A. Gisbrecht\,$^{1}$ \and B. Hammer\,$^{1}$}
\date{$^{1}$Univ. of Bielefeld, CITEC Center of Excellence,\\ Universit\"atsstrasse 21-23, 33615 Bielefeld, Germany\\ 01. September 2011\\[0.5cm]
Technical report follow of the Dagstuhl Seminar {\bf 11341}\\[0.3cm]
{\bf Learning in the context of very high dimensional data}\\
{21.08.11 - 26.08.11}\\[0.3cm]
Organizer: Michael Biehl (Univ. of Groningen, NL), Barbara Hammer (Univ. Bielefeld, DE), Erzsébet Merényi (Rice Univ., US),\\
Alessandro Sperduti (Univ. of Padova, IT), Thomas Villmann (Univ. of Applied Sc. Mittweida, DE)}
\begin{document}

\maketitle

\begin{abstract}
{\bf Motivation:} The analysis of physiological processes over time is becoming increasingly important. The measurements
are often given by spectrometric or gene expression profiles over time with only few time points 
but a large number of measured variables. The analysis
of such temporal sequences is challenging and only few methods have
been proposed. The information can be encoded time independent, by means of classical expression
differences for a single time point or in expression profiles over time. Available methods
are limited to unsupervised and semi-supervised settings. The predictive variables
can be identified only by means of wrapper or post-processing techniques. This is complicated due to the small number
of samples for such studies. Here, we present a supervised learning approach, termed 
\emph{Supervised Topographic Mapping Through Time} (SGTM-TT). It learns a supervised mapping of the
temporal sequences onto a low dimensional grid. We utilize a hidden markov model (HMM) to
account for the time domain and relevance learning to identify the relevant feature dimensions
most predictive over time. The learned mapping can be used to visualize the temporal sequences and
to predict the class of a new sequence. The relevance learning permits the identification of discriminating
masses or gen expressions and prunes dimensions which are unnecessary for the classification task or
encode mainly noise. In this way we obtain a very efficient learning system for temporal sequences.

{\bf Results:} The results indicate that using simultaneous supervised learning and metric adaptation significantly improves
the prediction accuracy for synthetically and real life data in comparison to the standard techniques. The 
discriminating features, identified by relevance learning, compare favorably with the results of alternative
methods. Our method permits the visualization of the data on a low dimensional grid, highlighting
the observed temporal structure.  

{\bf Contact:} \href{fschleif@techfak.uni-bielefeld.de}{fschleif@techfak.uni-bielefeld.de}
\end{abstract}


\section{Introduction}
The analysis of high-dimensional, short time series, or temporal sequences is a challenging task.
On the one hand side the data are not any longer identical and independent distributed (i.i.d) due to the
time constraint, on the other hand the dimensionality of the data is large, complicating
the learning of a predictive model. Standard time series methods like auto-regressive moving average (ARMA) 
or extensions thereof (see e.g. \cite{Hamilton1994a}) are in general not applicable due to the
limited number of time points and the large dimensionality of the data. Some methods have been proposed to
model this type of data. In \cite{neucom05strickertmsom} an unsupervised projection techniques was
proposed employing a so called temporal context. The temporal data have been processed by a kind of Self Organizing Map
(SOM) \cite{kohonen95a} but the learning was modified such that it depends on the the current temporal
context. A further unsupervised proposal has been made in 
\cite{DBLP:journals/nn/OlierV08} using the Generative Topographic Mapping Through Time (GTM-TT) (\cite{Bishop97gtmthrough}).
Some new hidden variables were introduced to account for the relevance of the different feature dimensions,
to accounts, in a non-discriminative manner, for the explained variance in the data over time.
A supervised two-class method solely based on hidden markov models was proposed in \cite{DBLP:conf/ismb/LinKB08}.
It models the two different data distribution by independent HMMs and evaluates the generated models to obtain a
ranking of the input dimensions. Subsequently the model was improved by selecting a set of features using
a wrapper strategy. In \cite{DBLP:journals/bioinformatics/CostaSHS09} a similar approach was proposed but in
a semi-supervised scenario, introducing classwise constraints in the hidden markov model. The importance of
the individual features was determined using a complex post processing procedure. Another supervised method using all 
features, based on Support Vector Machine (SVM) and a Kalman filter was proposed in \cite{DBLP:conf/psb/BorgwardtVK06}.

While the first two approaches have been found to be very effective for unsupervised analysis, the last mentioned methods 
focus on supervised and semi-supervised analysis. The results in \cite{DBLP:conf/ismb/LinKB08} are very promising, 
with $85\%$ prediction accuracy on a real life multiple sclerosis data (MS) set, but make strong pre-assumptions about the underlying
HMM structure. Also, it is proposed for two class scenarios, only. The approach in \cite{DBLP:conf/psb/BorgwardtVK06}
improved this result by $2-5\%$ but in a black box scenario, without additional feature selection. 
The approach in \cite{DBLP:journals/bioinformatics/CostaSHS09} is evaluated also with respect to the results of \cite{DBLP:conf/ismb/LinKB08} 
achieving improved performance for the same MS data sets. There is still ongoing work of research in this field, reflecting the high demand for effective methods dealing with this
type of data. The application field is not limited to the bio-medical domain as considered in \cite{DBLP:conf/ismb/LinKB08,
DBLP:journals/bioinformatics/CostaSHS09,Schliep2011a} but covers a broader field of applications also in industry and
geo-science as reflected in \cite{DBLP:journals/nn/OlierV08,neucom05strickertmsom}.

The identification of the relevant input dimensions of a temporal sequence is very important as outlined in \cite{DBLP:journals/nn/OlierV08,DBLP:conf/ismb/LinKB08}
to obtain better understanding of the data, to reduce the processing complexity and to improve the overall prediction accuracy. 
As already motivated by some of the prior references, prototype methods (see e.g.  \cite{kohonen95a}) have been found to be very effective for the analysis of high dimensional 
data also to analyze temporal sequences. In \cite{Bishop97gtmthrough}, the \emph{Generative Topographic Mapping -
through time} (GTM-TT), an unsupervised prototype based method for the topographic projection of high-dimensional, 
temporal sequences was proposed. GTM-TT learns a hidden markov model (HMM) of a data generating process and represents
the data by a prototype based representation in time and space. Like in ordinary prototype methods the GTM-TT
approximates the data distribution by a vector quantization of the data space. The temporal dependence between
the prototype is modeled by an appropriate HMM. Additionally the prototypes are assigned to a fixed grid representation
or lattice, which permits, provided the topology is preserved (see \cite{villmann97bRef}), the easy visualization and interpretation of the data trajectory 
in a low dimensional space. In this contribution 
we extend the GTM-TT to a supervised method and integrate relevance learning to identify the relevant dimensions over
time. Then we will briefly review Generative Topographic Mapping (GTM) and Generative Topographic Mapping Through Time.
Subsequently, we outline our method and apply and discuss it for different experimental data. The paper is closed with 
links to further extensions and open questions.

\section{Approach and Methods}

\subsection{Generative Topographic Mapping}
The Generative Topographic Mapping (GTM) as introduced in \cite{DBLP:journals/neco/BishopSW98} models data $\mathbf{x}\in\mathbb{R}^D$
by means of a mixture of Gaussians which is induced by a lattice of points $\mathbf{w}$ 
in a low dimensional latent space which can be used for visualization.
\begin{figure}
\centering
	\includegraphics[width=0.9\columnwidth,height=0.3\textheight]{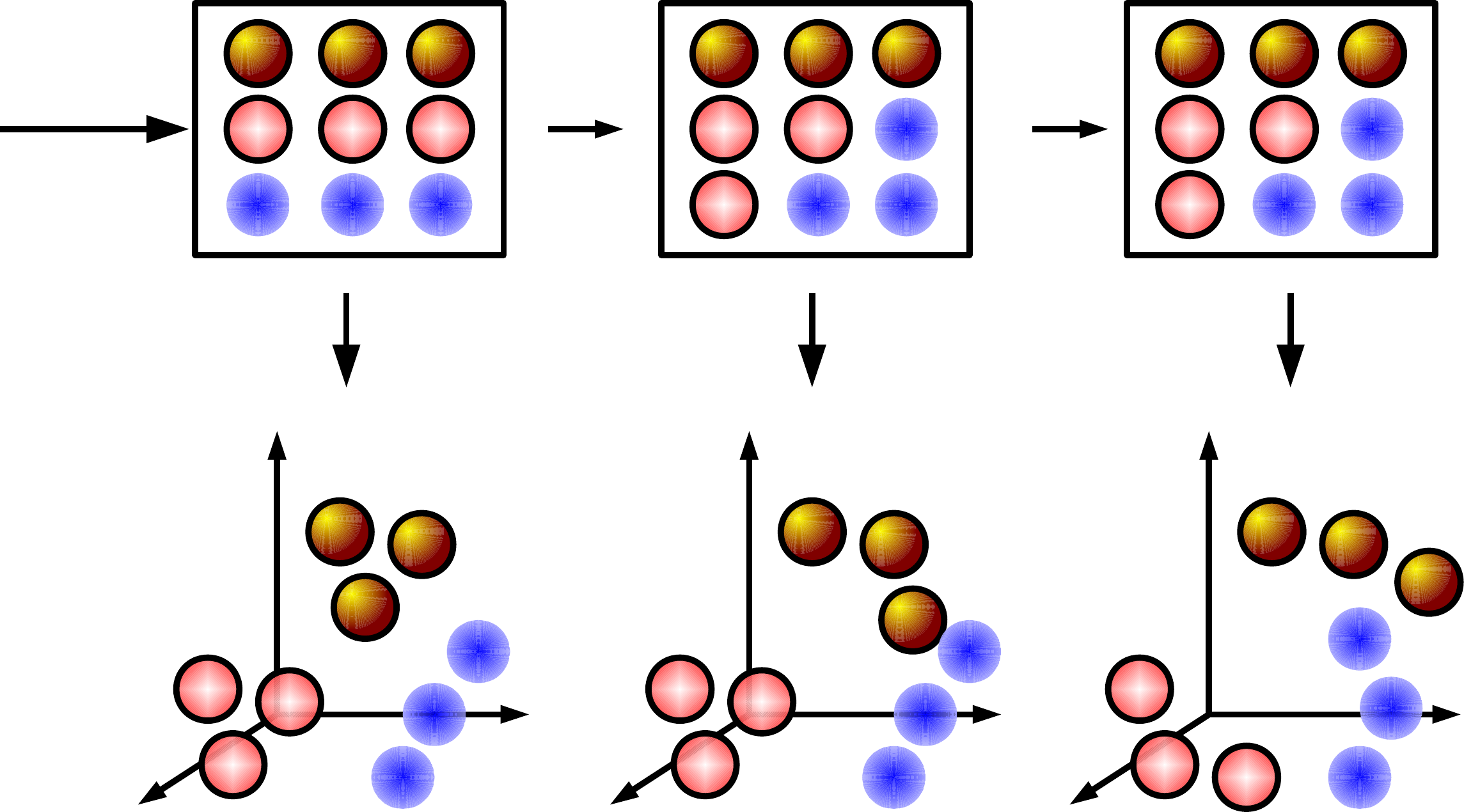}
	\caption{\label{fig:gtmtt}GTM-TT consisting of a HMM in which the hidden
	states are given by the latent points of the GTM model. The emission probabilities are governed 
	by the GTM mixture distribution \cite{Bishop97gtmthrough}. The different data distributions,
	exemplified in 3D (bottom) and indicated by the color/shading are mapped to the 2D grid (top). 
	Here we consider $9$ hidden states on a $3 \times 3$ grid. 
	The data distribution may change over time and hence also the mapping of the GTM is effected over time,
	assuming smooth transitions.}
\end{figure}

The lattice points are mapped via 
$\mathbf{w}\mapsto \mathbf{t}=y(\mathbf{w},\mathbf{W})$ to the data space,
where the function is parametrized by $\mathbf{W}$; one can, for example, pick a generalized linear regression model 
based on Gaussian base functions
\begin{equation}
	y:\mathbf{w}\mapsto
	\Phi(\mathbf{w})\cdot \mathbf{W}
	\label{regression}
\end{equation}
where the base functions $\Phi$ are equally spaced Gaussians.
The high-dimensional points $y$ are so called prototypes of the original data space, representing
a larger set of points, they are responsible for, as measured by Eq. \eqref{responsibility}.
They can be directly inspected and permit to summarize the data.
Every latent point induces a Gaussian
\begin{equation}
	p(\mathbf{x} | \mathbf{w}, \mathbf{W}, \beta) = \left(\frac{\beta}{2 \pi} \right)^{D/2} 
	\exp \left(- \frac{\beta}{2} \|\mathbf{x}-y(\mathbf{w}, \mathbf{W})\|^2 \right)\label{bell}
\end{equation}
with variance $\beta^{-1}$, which gives the data distribution as mixture of $K$ modes
\begin{equation}
	p(\mathbf{x}| \mathbf{W}, \beta) =\sum_{k=1}^K p(\mathbf{w}^k)p(\mathbf{x} | \mathbf{w}^k, \mathbf{W}, \beta)
	\label{mixture}
\end{equation}
where, usually, $p(\mathbf{w}^k)$ is taken as Dirac distributions of the prototypes. Training of GTM optimizes the data log-likelihood
\begin{equation}
	\ln \left( \prod_{n=1}^N \left( \sum_{k=1}^K p(\mathbf{w}^k)p(\mathbf{x}^n | \mathbf{w}^k, \mathbf{W}, \beta) \right) \right)
\end{equation}
by means of an expectation maximization (EM) approach with respect to the parameters $\mathbf{W}$ and $\beta$.
In the E step, the responsibility of mixture component $k$ for point $n$ is determined as
\begin{equation}
  r^{kn}=p(\mathbf{w}^k | \mathbf{x}^n, \mathbf{W}, \beta)=
  \frac{p(\mathbf{x}^n | \mathbf{w}^k, \mathbf{W}, \beta) p(\mathbf{w}^k)}
  {\sum_{k'} p(\mathbf{x}^n | \mathbf{w}^{k'}, \mathbf{W}, \beta) p(\mathbf{w}^{k'})}
  \label{responsibility}
\end{equation}
In the M step, the weights $\mathbf{W}$ are determined solving the equality
\begin{equation}
	\mathbf{\Phi}^T\mathbf{G}_{\mathrm{old}}\mathbf{\Phi}\mathbf{W}_{\mathrm{new}}^T
	=\mathbf{\Phi}^T\mathbf{R}_{\mathrm{old}}\mathbf{X}
	\label{weights}
\end{equation}
where $\mathbf{\Phi}$ refers to the matrix of base functions $\Phi$ evaluated at points $\mathbf{w}^k$,
$\mathbf{X}$ to the data points, $\mathbf{R}$ to the responsibilities, and  $\mathbf{G}$ is a diagonal matrix with accumulated responsibilities
$G_{nn}=\sum_kr^{kn}(\mathbf{W},\beta)$. The variance can be computed by solving
\begin{equation}
	\frac{1}{\beta_{\mathrm{new}}}=\frac{1}{ND}\sum_{k,n}
	r^{kn}(\mathbf{W}_{\mathrm{old}},\beta_{\mathrm{old}})
	\|{\Phi}(\mathbf{w}^k)\mathbf{W}_{\mathrm{new}}-\mathbf{x}^n\|^2
\label{variance}
\end{equation}
where $D$ is the data dimensionality and $N$ the number of data points.

\subsection{Relevance learning}
The principle of relevance learning has been introduced in \cite{nn02hammer} as a particularly simple and efficient method
to adapt the metric of prototype based classifiers according to the given situation at hand. It takes into account a relevance
scheme of the data dimensions by substituting the squared Euclidean metric by the weighted form
\begin{equation}\label{eq:weuc}
	d_{\boldsymbol{\lambda}}(\mathbf{x},\mathbf{t})=
	\sum_{d=1}^D \lambda_d^2(x_d-t_d)^2\,.
\end{equation}
The principle is extended in \cite{ieeetnn10schetal,nc09schetal} to the more general metric form
\begin{equation}
	d_{\boldsymbol{\Omega}}(\mathbf{x},\mathbf{t})=
	(\mathbf{x}-\mathbf{t})^T \boldsymbol{\Omega}^T \boldsymbol{\Omega} (\mathbf{x}-\mathbf{t})
	\label{matrix}
\end{equation}
Using a square matrix $\boldsymbol{\Omega}$, a positive semi-definite matrix which gives rise to a valid pseudo-metric is achieved this way.
In \cite{ieeetnn10schetal,nc09schetal}, these metrics are considered  in local and global form, i.e.\ the adaptive metric parameters can be identical for the full model,
or they can be attached to every prototype present in the model. 
Relevance learning for GTM has been already introduced in \cite{neucom11gisham}
for i.i.d. data. In case of temporal sequences some modification of the original principle are necessary and also the supervision will be handled
differently as pointed out subsequently. First however we review the GTM through time as described in \cite{Bishop97gtmthrough,DBLP:conf/ijcnn/OlierV08} which is the 
basic method to process i.i.d. data in our approach.

\subsection{Generative Topographic Mapping Through-Time}
The GTM through time (GTM-TT) has been introduced in \cite{Bishop97gtmthrough}. For data vectors $\mathbf{x}$ which have the form of a time series
the vectors are no longer independent. Nearby timepoints are likely to be correlated. As pointed out in \cite{Bishop97gtmthrough} such effects
can be captured using Hidden Markov Models (HMM). Accordingly in \cite{Bishop97gtmthrough} the GTM is equipped by a HMM, constructing a kind of
a topology-constrained HMM

The structure of the GTM-TT is shown in Figure \ref{fig:gtmtt}. Assuming a sequence length $T$, of hidden states $Z=\{z_1,\hdots,z_n,\hdots z_T\}$
and the observed multidimensional time series $X=\{x^1,x^2,\hdots,x^n,\hdots x^T\}$, the probability of the observations is given by
\begin{equation}
	p(X) = \sum_{\text{all sequences } Z} p(Z,X)
\end{equation}
where $p(Z,X)$ defines the complete data likelihood as in HMM models \cite{DBLP:journals/neco/BishopSW98} taking the following form:
\begin{equation}
	p(Z,X) = p(z_1) \prod_{n=2}^T p(z_n|z_{n-1}) \prod_{n=1}^T p(x^n|z_n)
\end{equation}
So it consists of the initial state probability, the transition probability between two hidden states, capturing the temporal dependence, and the probability to observe
a specific sequence in a given state also known as emission probability (covered by Eq. \eqref{bell}). The model parameters are $\Theta = (\pi, A, W, \beta)$ 
where $\pi=\{\pi_j\}:\pi_j:= p(z_1=j)$ are the initial state probabilities. $A=\{a_{ij}\}: a_{ij}=p(z_n=j|z_{n-1} = i)$ are the transition state probabilities,
and $\{W,\beta\}$ are given by Eq. \eqref{weights}. Again
we control the gaussians by a common invariance $\beta$. As in HMM the above likelihood can be efficiently calculated using the \emph{forward backward procedure} \cite{BaumWelch2003}.
The probability being in state $\mathbf{w}_k$ at time $n$, given the observation sequence and the model, also known as responsibility $r^{kn}$ is calculated as:
\begin{equation}
	r^{kn} = p(z_n = \mathbf{w}^k | X,\Theta) = \frac{A_{kn}B_{kn}}{p(X|\Theta)}
\end{equation}
The forward variable $A_{kn}$ is the joint probability of the past sequences $\{x^1,\hdots, x^n\}$ and the state $z_n=\mathbf{w}^k$,
i.e. $A_{kn} = p(\{x^1,\hdots, x^n\},z_n=\mathbf{w}^k|\Theta)$, given by the following recursive equation:
\begin{equation}\label{eq:forward}
	A_{kn} = \left ( \sum_{i=1}^K A_{i,n-1} p_{i,k} \right ) p_k (x^n)
\end{equation}
where $A_{k,1}=\pi_k p_k(x^1)$. The backward variable $B_{kn}$ which is the probability of the future sequence $x^{n+1},x^{n+2},\hdots,x^N$
given the hidden state $z_n=\mathbf{w}^k$, i.e. $B_{kn} = p(\{x^{n+1},x^{n+2},\hdots,x^N\}|z_n=\mathbf{w}^k,\Theta)$ is calculated using the
following recursive equation:
\begin{equation}
	B_{kn} = \sum_{i=1}^K p_{i,k} p_i(x^{n+1}) B_{i,n+1}
\end{equation}
where $B_{k,T} = 1$. The whole parameter estimation can be accomplished by a maximum likelihood optimization using the EM algorithm as sketched above.
Details can be found in \cite{gtmttphd}.

\subsection{Supervised GTM-TT}
Assume that data point $X$ is equipped with label information $l$ which is element of a finite set of different labels $L$,
e.g. $L=\{0,1\}$.  Lets assume we have only two labels \footnote{An extension to multiple labels is straight forward.}. 
The data are divided into two groups, according to the labeling and we train one separate GTM-TT per group. To keep the models
comparable, the $\beta$ update for the models is linked to each other and optimized in the outer loop. 
The parameters $\mathbf{W}$ are determined for each model individually leading to $\mathbf{W}_0$ and $\mathbf{W}_1$. 
We will further assume that the grid structure is common for both models. 
The learning procedure is thus similar to GTM-TT and depicted in Figure \ref{alg:sgtmtt}.

\begin{figure}
\centering
	\includegraphics[width=0.9\columnwidth,height=0.3\textheight]{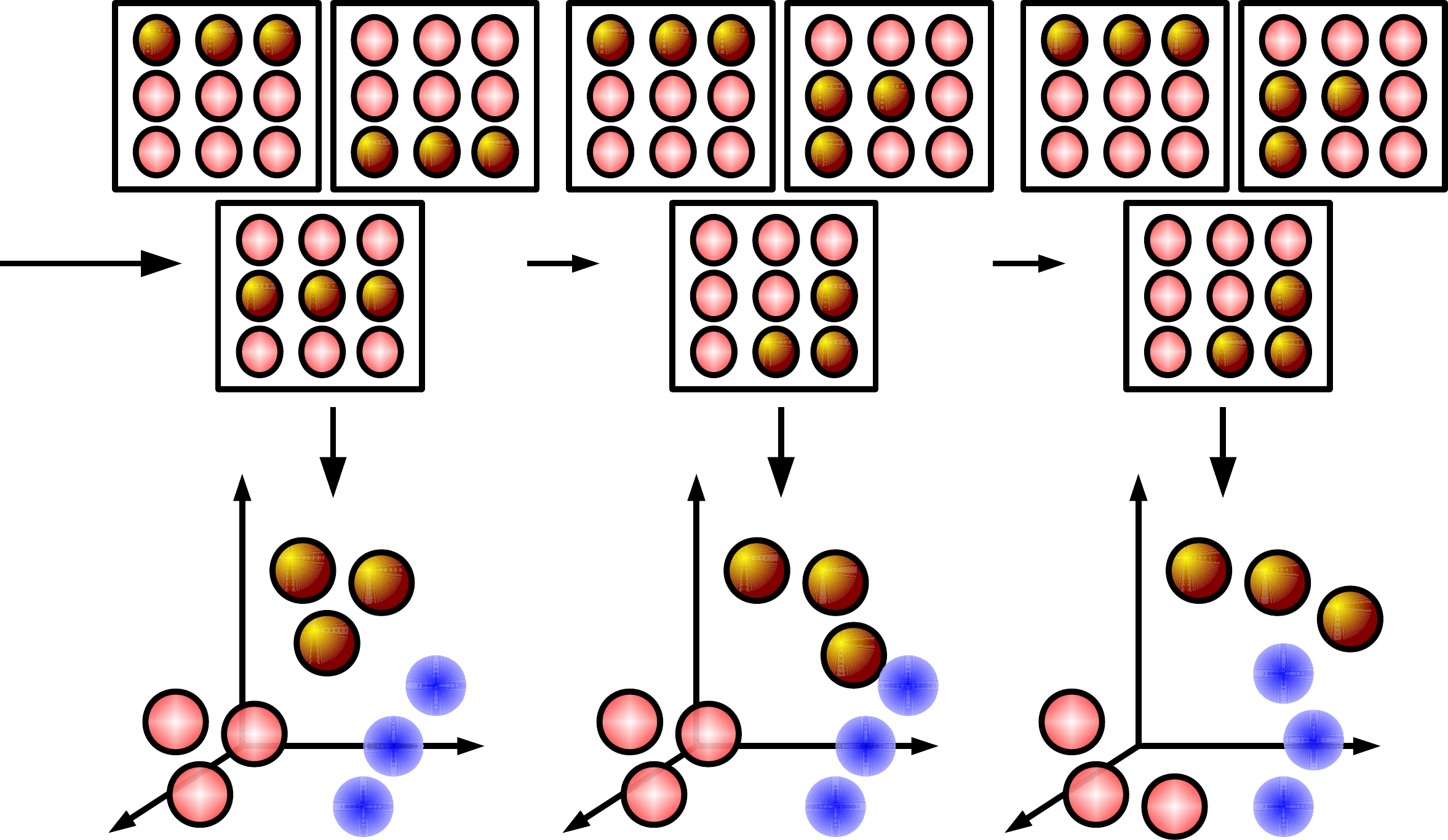}
	\caption{\label{fig:sgtmtt}Illustration of the SGTM-TT. It consists of multiple GTM-TT models. It behaves similar to the regular
        GTM-TT but the training is classwise and the $\beta$ parameter is common over the different models.
        The different classwise models (top) are used to represent
	the data distribution (bottom) over time (from left to right).}
\end{figure}

\begin{center}
\begin{algorithm}[ht]
	\footnotesize
 \begin{algorithmic}[1]
   \Function{Supervised GTM-TT}{$X$,$L$,$K$}
       \State \text{[Xn,Pars    ] = normalize(X)}
       \State \text{[X1,X2,L1,L2] = splitdata(Xn,L)}
       \State [$M_0$, $M_1$] = \text{\textbf{init} both GTM-TT models}
       \Repeat
               \State \text{\textbf{call} \textit{train\_single\_step} for $M_0$, $M_1$}
               \State \text{\textbf{call} \textit{convergence\_check}  for $M_0$, $M_1$}
               \State \text{\textbf{call} \textit{optimize\_beta}      for $M_0$, $M_1$}
	       \State $\bar{\beta}=$ \text{calculate mean of the } $\beta$
	       \State \text{\textbf{call} \textit{update\_beta($M_0$, $M_1$,$\bar{\beta}$})}
       \Until{convergence is true for both models}       
   \EndFunction
 \end{algorithmic}
 \caption{Pseudocode of supervised GTM-TT}
 \label{alg:sgtmtt}
\end{algorithm}
\end{center}
We denote the obtained model as Supervised GTM-TT (SGTM-TT) and the submodels as $M_0$ and $M_1$.
The concept of the SGTM-TT is depicted schematically in Figure \ref{fig:sgtmtt}.

\subsection{Classification using SGTM-TT}
To classify new data points with the SGTM-TT model different approaches can be taken.  The simplest one is to make direct
use of the samplewise likelihoods considering the class wise models. In that case a new point is assigned to the model with maximal likelihood
considering one model against the rest. A more interesting approach is to combine the performance of the generative SGTM-TT model 
with a discriminative approach like the SVM \cite{Vapnik1995a}. Again we use the likelihood values from the forward procedure \eqref{eq:forward} of the SGTM-TT
and define a kernel as follows:
\begin{eqnarray}
	Lik_j^l    &=& \sum_{i=1}^K A_{i,j} \text{ for a series $j$ and a sub-model } l \\
	K(X_j,X_k) &=& \sum_{l=1:\#L} Lik_j^l \cdot Lik_k^l \; \text{ with equal prior}
\end{eqnarray}
Hence the kernel $K$ is based on a kernel function of inner-products in a one dimensional feature space of the likelihood-values. 
In the following we will make use of this approach employing a standard SVM implementation.

\subsection{Relevance learning for SGTM-TT}
Relevance learning for GTM has been introduced in \cite{neucom11gisham}, as the Relevance GTM (R-GTM). The basic idea for Relevance GTM is
to introduce an adaptive metric for the GTM. The original Euclidean metric is replaced by a parametric distance like the weighted
Euclidean metric \eqref{eq:weuc}. After each GTM training step the prototypes are post-labeled according to its responsibilities, 
employing the labeling $L$ of the datapoints. Subsequently the metric parameters of the distance are adapted according to an
optimization criterion. In the article of \cite{neucom11gisham} different cost functions E where suggested. 

The data of the GTM-TT are not any longer i.i.d. and, as mentioned before we observe a sequence of states $Z$ for a given time series $X$.
In the SGTM-TT we know the labeling of the prototypes, assuming constant labels over time, due to the split of the learning problem
according to the data labeling. Further, using a common metric and common $\beta$ parameters the prototypes $Y$ exist still in the same 
common dataspace. Relevance learning can now be done in the same way as for R-GTM. This however is often not useful because the original
relevance learning ignores the time domain. If data separation is observed over time and not for a single time point the R-GTM approach will
fail. For temporal sequences we may also be interested on two views of relevance, namely relevant, or separating input \emph{dimensions} $x_i$ but
also relevant \emph{time points} in a temporal sequence $x$. Taking this problem into account we consider two distance measures, one for the time
domain, denoted as $d^t$ and one for the time-independent data space $d^{\lambda}$. A parametrization of $d^t$ can be used to account for
the relevance of specific time points, e.g to prune out time points which are irrelevant for the representation of the data in a discriminative
manner. Parameters on $d^\lambda$ can be used to identify discriminating feature dimensions, e.g. to prune out noisy dimensions. Subsequently,
we provide a distance measure which can be used for $d^t$ and a specific form for $d^\lambda$. For simplicity we will use a simple \emph{global},
\emph{diagonal} metric learning scheme in the experiments.

SGTM-TT provides a probabilistic prediction of the internal representation $\hat{\mathbf{x}}$ of a time series $\mathbf{x}$ 
considering the two GTM-TT models, we obtain one reconstruction each:
\begin{eqnarray*}
	\hat{\mathbf{x}^n}^l_i &=& Y_l(\underset{k}{\arg\max} \left( r^{kn} \right),i) \; \forall i \in [1,D]\\
	& & \text{with } l \in \{0,1\} 
\end{eqnarray*}
Now, two distances are calculated over time for each point and each dimension $i$: 
$d^t(\hat{\mathbf{x}^n}^0_i,\mathbf{x}^n_i)$, $d^t(\hat{\mathbf{x}^n}^1_i,\mathbf{x}^n_i)$. 
Using one of the suggested cost functions in the paper of \cite{neucom11gisham} we
can calculate the relevance of the individual dimensions for the separation between
the two reconstructions per point and hence between the different models. 

Like for R-GTM the metric adaptation is done by an appropriate optimization scheme on the cost functions,
here we will use stochastic gradient descend, with a fixed learning rate $\epsilon=0.1$. 
To avoid convergence to trivial optima such as zero we pose constraints on the metric parameters
of the form $\|\boldsymbol{\lambda}\|=1$ or $\mathrm{trace}(\boldsymbol{\Omega}^T \boldsymbol{\Omega})^2=1$, for matrix learning.
This is achieved by normalization of the values, i.e.\ after every gradient step,
$\boldsymbol{\lambda}$ is divided by its length, and $\boldsymbol{\Omega}$ is divided by the square
root of $\mathrm{trace}(\boldsymbol{\Omega}^T \boldsymbol{\Omega})$.

A pseudo code of the SGTM-TT with relevance learning is depicted in \ref{alg:sgtmttr}.
\begin{center}
\begin{algorithm}[ht]
	\footnotesize
 \begin{algorithmic}[1]
   \Function{SGTM-TT-R}{$X$,$L$,$K$}
       \State \text{[Xn,Pars    ] = normalize(X)}
       \State \text{[X1,X2,L1,L2] = splitdata(Xn,L)}
       \State \text{initialize the common metric}
       \State [$M_0$,$M_1$] = \text{\textbf{init} both GTM-TT models}
       \Repeat
               \State \text{\textbf{call} \textit{train\_single\_step} for each GTM-TT model}
               \State \text{\textbf{call} \textit{convergence\_check} for each GTM-TT model}
	       \If{$cycle > 10$} 
			\State \text{$\forall X, \forall i=1:D$ \textbf{call} \textit{reconstruct($X_i,M_0,M_1$)}}
			\State \text{$\forall X, \forall i=1:D$ \textbf{call} \textit{$d^t(M_0,\hat{x_i}_0,x_i)$}}
			\State \text{$\forall X, \forall i=1:D$ \textbf{call} \textit{$d^t(M_1,\hat{x_i}_1,x_i)$}}
			\State \text{$\forall X$ \textbf{call} \textit{calculate\_metric\_update}}
			\State \text{average the metric updates and normalize}
			\State \text{update the metric parameter annealed by $\epsilon$}
	       \EndIf
               \State \text{\textbf{call} \textit{optimize\_beta}     for each GTM-TT model}
	       \State $\bar{\beta}=$ \text{calculate mean of the } $\beta$
	       \State \text{\textbf{call} \textit{update\_beta(M1,M2,$\bar{\beta}$})}
       \Until{convergence is true for both models}       
   \EndFunction
 \end{algorithmic}
 \caption{Pseudocode of supervised GTM-TT with relevance learning}
 \label{alg:sgtmttr}
\end{algorithm}
\end{center}

Usually, we alternate between one EM step, one epoch of gradient descent, and normalization in our experiments
and start the metric learning after $10$ epochs of EM learning to allow a reasonable pre-positioning of the GTM-TT
in the dataspace. The metric learning is annealed by $\epsilon$. Since EM optimization is much faster than gradient descent, 
this way, we can enforce that the metric parameters are adapted on a slower time scale.
Hence we can assume an approximately constant metric for the EM optimization, i.e.\ the EM scheme optimizes the likelihood as before.
Metric adaptation takes place considering quasi stationary states of the GTM solution due to the slower time scale.
The call of \textit{train\_single\_step} is a regular EM optimization step of the GTM-TT but without the adaptation of the parameter
$\beta$ which is postponed to allow a linking between the two GTM-TT models included in the SGTM-TT. 

Now, we briefly review a concrete cost function $E$ of the relevance GTM for the metric adaptation as already introduced in 
\cite{neucom11gisham} but account for the alternative distance calculations mentioned before.

\subsubsection*{Cost function - Generalized Relevance GTM (GRGTM)}
Metric parameters have the form $\boldsymbol{\lambda}$ or $\boldsymbol{\lambda}^k$ for a diagonal metric
\eqref{eq:weuc} and $\boldsymbol{\Omega}$ or $\boldsymbol{\Omega}^k$ for a full matrix (\ref{matrix}),
depending on whether a local or global scheme is considered.
In the following, we define the general parameter
$\Theta^k$ which can be chosen as one of these four possibilities
depending on the given setting. Thereby, we can assume that $\Theta^k$ can be realized by a matrix which
has diagonal form (for relevance learning) or full matrix form (for matrix updates).

The cost function of generalized relevance GTM is taken from generalized relevance learning vector 
quantization (GRLVQ), which can be interpreted as maximizing  the hypothesis margin of a prototype based
classification scheme \cite{nn02hammer,ieeetnn10schetal}. The cost function has the form 

\begin{equation}
E(\Theta)= \sum_nE_n(\Theta)= \sum_{n} \operatorname{sgd} \left(
				\frac{d_{\Theta^+}(\mathbf{x}^n,\hat{\mathbf{x}^n}^+)-
                              d_{\Theta^-}(\mathbf{x}^n,\hat{\mathbf{x}^n}^-)}
                             {d_{\Theta^+}(\mathbf{x}^n,\hat{\mathbf{x}^n}^+)+
                             d_{\Theta^-}(\mathbf{x}^n, \hat{\mathbf{x}^n}^-)}
							\right)
\end{equation}
where $\operatorname{sgd}(x)=(1+\exp(-x))^{-1}$, $\hat{\mathbf{x}^n}^\pm$ is the reconstruction of $\mathbf{x}^n$ over time
using the model $M_0$ or $M_1$ depending on the label of $\mathbf{x}$, $+$ indicates the model with the same level $-$ 
the model with a different label or the model for the remaining data.

The adaptation formulas can be derived thereof by taking the derivatives with respect to the metric parameter.
Depending on the form of the metric, the derivative of the metric is simple
\begin{equation}
\frac{\partial d_{\boldsymbol{\lambda}}(\mathbf{x},\hat{\mathbf{x}^n})}
{\partial \lambda_i}=
2\lambda_i d^t(x_i,\hat{x^n}_i)^2
\end{equation}
for a diagonal metric and
\begin{equation}
\frac{\partial d_{\boldsymbol{\Omega}}(\mathbf{x},\hat{\mathbf{x}^n})}
{\partial \Omega_{ij}}=
2d^t(x_j,\hat{x^n}_j)\sum_d\Omega_{id}d^t(x_d,\hat{x^n}_d)
\end{equation}
for a full matrix.

For simplicity, we denote the respective squared distances to
the closest correct and wrong model, respectively, by $d^+=d_{\Theta^+}(\mathbf{x}^n,\hat{\mathbf{x}}^+)$ and $d^-=d_{\Theta^-}(\mathbf{x}^n,\hat{\mathbf{x}}^-)$.
The term $\operatorname{sgd}'$ is a shorthand notation for  $\operatorname{sgd}'((d^+-d^-)/(d^++d^-))$.
Given a data point $\mathbf{x}^n$ the derivative of the corresponding summand of cost function $E$ with respect to metric parameters yields 
\begin{equation}
	\frac{\partial{E_n}}{\partial \Theta^+}= 2\operatorname{sgd}'\cdot \frac{d^-}{(d^++d^-)^2}\cdot\frac{\partial d^+}{\partial\Theta^+}
\end{equation}
for the parameters of the closest correct prototype and
\begin{equation}
	\frac{\partial{E_n}}{\partial \Theta^-}= -2\operatorname{sgd}'\cdot \frac{d^+}{(d^++d^-)^2}\cdot\frac{\partial d^-}{\partial\Theta^-}
\end{equation}
for the parameters attached to the closest wrong model. All other parameters are not affected.
As pointed out before we choose only a global metric such that the update corresponds to the sum of these two
derivatives.

\subsubsection*{Distance measure for functional data}\label{subsec:measures}
\begin{figure}
\centering
\subfigure[Two functions: Euc = $L^p$-norm]{\includegraphics[width=0.48\columnwidth]{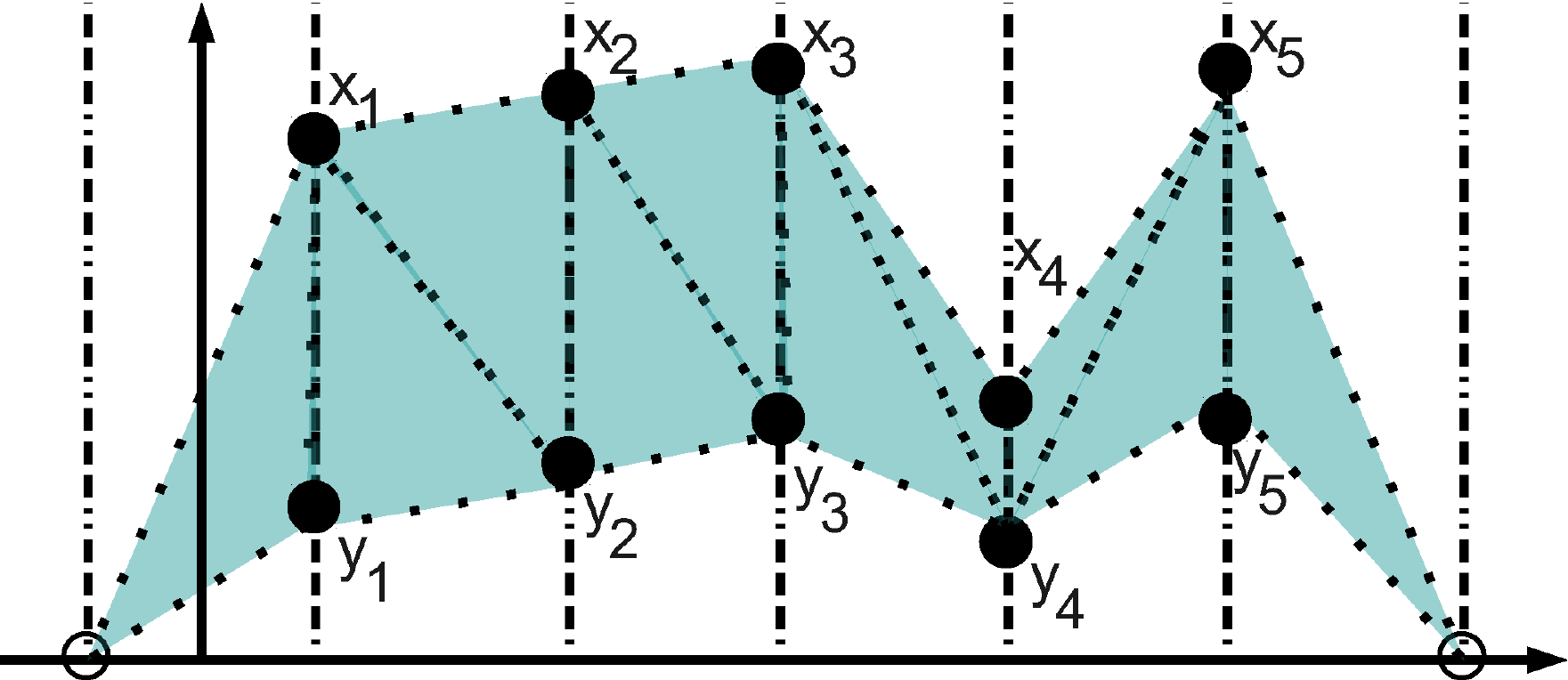}}\quad
\subfigure[Two functions: Euc $\ne$ $L^p$-norm]{\includegraphics[width=0.48\columnwidth]{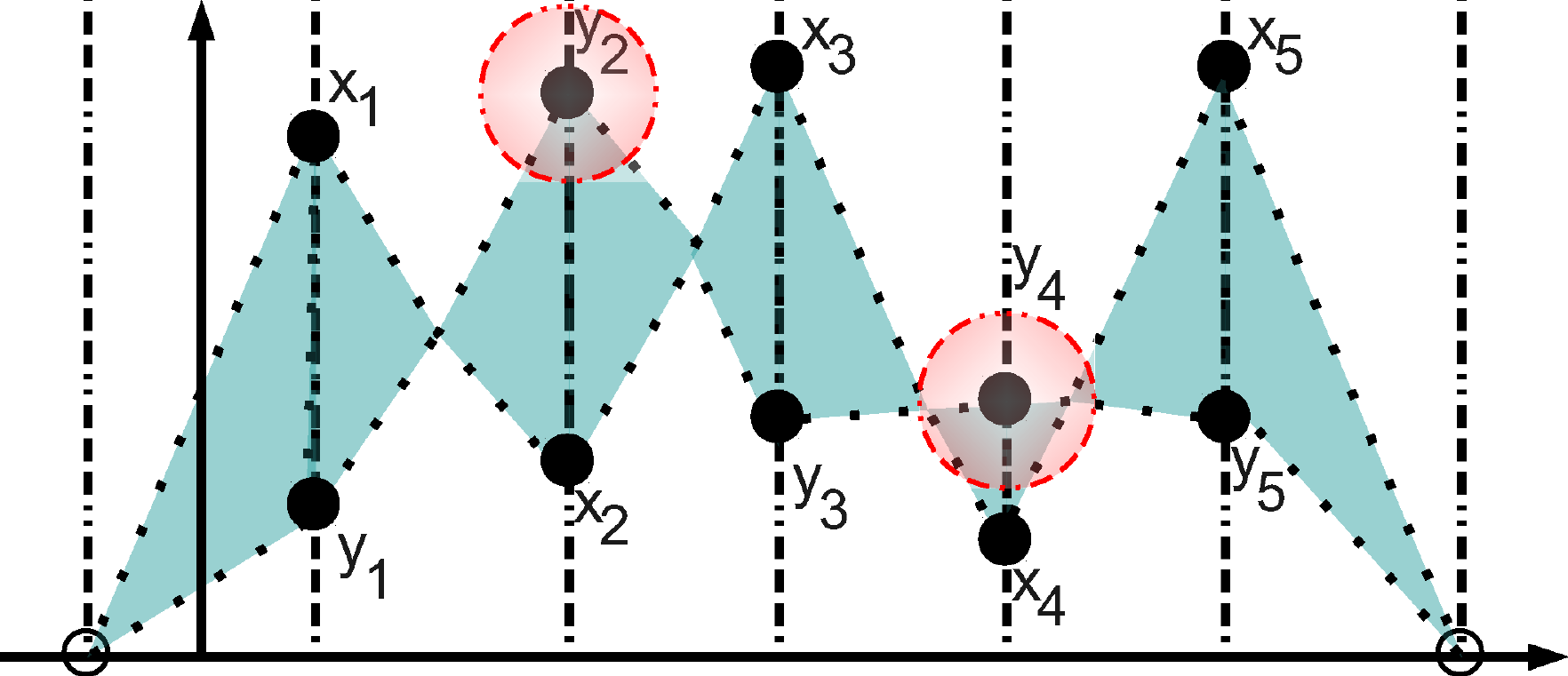}}
\caption{\label{fig:lpnorm}Illustration of the $L^p$-norm. Plot (a) indicates the case in which the distance between two functions is equal, both for Euclidean or $L^p$-norm. 
In plot (b) parts of the functions are interchanging (crossing). The distance using Euc is still the same as in plot (a) but for the $L^p$-norm the distance is changed, giving a more 
realistic measure of the distance of the two functions.}
\end{figure}

Here we consider a functional distance measure as an extension of the $L^p$ norm proposed in (\cite{LeeVerleysen2005a}) subsequently denoted as
(FUNC). The functional distance measure has the advantage of taking the functional nature of the data into account, or in our case the dependence
over time, which also constitutes a function $f(t)$, with potentially discrete arguments $t$. It has been already successfully used for the analysis
of biomedical data as shown in \cite{Schleif2011b}.
The standard Euclidean distance considers the individual features of a signal independent, so that a change
in the order of the positions does not affect the calculated distance. However, the measurement points over time 
are not independent, so that a distance taking this aspect into account can be considered to be more appropriate for this type of data.
Lee proposed a distance measure taking the functional structure 
into account by involving the previous and next values of a signal $v_i$ in the $i$-th term of the sum, instead
of $v_i$ alone. Assuming a constant sampling period $\tau$, the proposed norm (FUNC) is: 
\begin{equation}\label{eq:lee_metric}
	\mathcal{L}_{p}^{fc}\left( \mathbf{v}\right) =\left( \sum_{k=1}^{D}\left(
	A_{k}\left( \mathbf{v}\right) +B_{k}\left( \mathbf{v}\right) \right)
	^{p}\right) ^{\frac{1}{p}}
\end{equation}
with
\begin{eqnarray}
	A_{k}\left( \mathbf{v}\right) = %
	\begin{cases}
		\frac{\tau }{2} \vert v_{k} \vert &  \text{if }0\leq v_{k}v_{k-1}\\ 
		\frac{\tau }{2} \frac{v_{k}^{2}}{\vert v_{k}\vert +\vert v_{k-1}\vert } & \text{if }0>v_{k}v_{k-1}%
	\end{cases}
\\
  B_{k}\left( \mathbf{v}\right) = %
	\begin{cases}
		\frac{\tau }{2} \vert v_{k} \vert & \text{if }0\leq v_{k}v_{k+1} \\
	  \frac{\tau }{2} \frac{v_{k}^{2}}{\vert v_{k} \vert + \vert v_{k+1} \vert } &  \text{if }0>v_{k}v_{k+1}
	\end{cases}	  
\end{eqnarray}
representing the triangles on the left and right sides of $v_i$ and $D$ being the data dimensionality. 
For the data considered in this paper $v$ is a time series or a prototype reconstruction.
As for $L_p$, the value of $p$ is assumed to be a positive integer. 
At the left and right extremes of the sequence, $v_0$ and $v_D$ are assumed
 to be equal to zero. The concept of the $L^p$-norm is shown in Figure \ref{fig:lpnorm}.
The calculation of this norm is slightly more complex than that of the standard Euclidean. 

\subsection{Data set description}
Subsequently we consider two data sets to evaluate our approach. 

\subsubsection{Simulated data sets}
The first one is a simulated two class scenario, proposed in the paper of \cite{DBLP:conf/ismb/LinKB08}. It consists of $100$ samples divided into two classes
of $50$ samples each. For each sample $100$ features have been generated with $8$ time points. Out of the $100$ features,
only $10$ where substantially differentiating between the classes. The generation mechanism behind the simulated data is to sample
the time series from a piecewise linear function. At a later step, sample-specific variation is included by shrinking and
expanding the curves. 

\subsubsection{Multiple sclerosis data}
The second data set is taken from \cite{10.1371/journal.pbio.0030002} (IBIS) in the prepared form, given in \cite{DBLP:journals/bioinformatics/CostaSHS09}.
The data are taken from a clinical study analyzing the response of multiple sclerosis (MS) patients to the treatment. Blood sample entrenched with
mono-nuclear cells from $52$ relapsing-remitting MS patients were obtained $0,3,6,7,12,18$ and $24$ months after initiation of IFN$\beta$ therapy.
This resulted on an average of $7$ measurements across the $2$ years. Expression profiles were obtained using one-step kinetic reverse-transcription
PCR over $70$ genes selected by the specialists to be potentially related to IFN$\beta$ treatment. Overall, $8\%$ of the measurements were missing
due to patients missing the appointments. After the two year endpoint, patients were classified as either good or bad responders, depending on
strict clinical criteria. Bad responders were defined as having suffered two or more relapses or having a confirmed increase of at least one
point on the expanded disability status scale (EDSS). A good responder was to have a suppression of relapses and not allowed to have an increase
on the EDSS. From the $52$ patients, $33$ were classified as good and $19$ as bad responders. A more detailed description of the data set
is available in the paper of \cite{10.1371/journal.pbio.0030002} and the supplemented material, therein.

\section{Results and Discussion}
For the simulated and the MS data set, we reanalyzed the classification accuracy of the SGTM-TT with $9$ hidden states and $4$ basis functions. The analysis was done
within a $4$ fold cross-validation with $5$ repetitions. We compared it with the general HMM classifier (HMM-Lin) and the discriminative HMM classifier 
(HHM-Disc-Lin) proposed in \cite{DBLP:conf/ismb/LinKB08}. We also included the results of \cite{10.1371/journal.pbio.0030002} who originally proposed the MS study, 
the analysis of \cite{DBLP:conf/psb/2006}, employing a Kalman Filter combined with an SVM approach and \cite{DBLP:journals/bioinformatics/CostaSHS09} proposing 
a semi-supervised analysis coupled with a wrapper and cut-off technique to identify discriminating features. 

\begin{figure*}
\centering
	\includegraphics[width=0.95\textwidth,height=0.3\textheight]{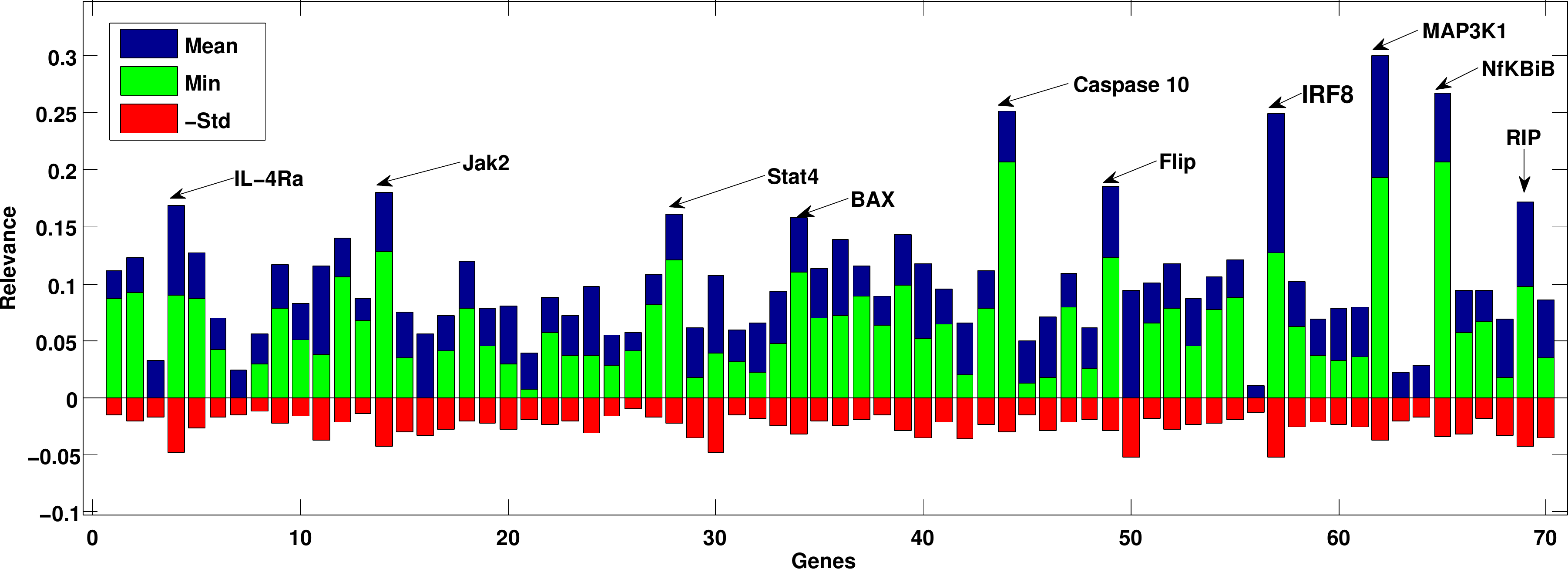}
\caption{\label{fig:msrelevance} Relevance profile as obtained using SGTM-TT with relevance learning. The plot shows
the average relevance (blue/dark), minimal relevance (green/bright) and the standard deviation of the relevance, flipped
to the negative part of the relevance axis. We observe that the standard-deviation is relatively small, hence the relevance
profiles of different runs are very stable. The most discriminative features (high-relevance), can in parts also be found in 
\cite{DBLP:journals/bioinformatics/CostaSHS09} but some additional features appear to be relevant, and our proposed set consists
of $7$ genes rather $17$ like in \cite{DBLP:journals/bioinformatics/CostaSHS09}}
\end{figure*}

\subsection{Simulated data}
We applied SGTM-TT with relevance learning for the simulated data set of \cite{DBLP:conf/ismb/LinKB08}. We observed an overall prediction accuracy of
$94 \pm 4$. The relevance profile identified all known $10$ features and effectively pruned out the remaining irrelevant data dimensions. 
Our results are slightly better than those reported in \cite{DBLP:conf/ismb/LinKB08} $(90\%)$ and by \cite{DBLP:journals/bioinformatics/CostaSHS09} $(92\%)$.

\subsection{Multiple sclerosis experiment}

\begin{table}[!t]
\begin{center}
\footnotesize
\begin{tabular*}{\columnwidth}{@{\extracolsep{\fill}}l|c|c}\hline
Method			& Number of genes & Test accuracy (\%) \\\hline\hline
SGTM-TT			& $70$            & $85.66\pm 8.3$       \\
SGTM-TT-R 		& $7$             & $93.43\pm 5.8$       \\
IBIS			& $3$             & $74.20$              \\
Kalman-SVM		& -               & $87.80$              \\
Lin-Best		& $ 7$            & $85.00$              \\
Costa-Best		& $17$            & $92.70\pm 6.1$       \\       
\end{tabular*}
\end{center}
\caption{\label{tab:results_ms} Prediction accuracies on the test data for different models using the MS data set. We observe improved predicition accuracy
employing feature selection. This is also true for SGTM-TT which improved by $\approx 6\%$ using relevance learning and the SVM classifier. Interestingly also the
prediction accuracy on the full data set, including all features and without relevance learning is quite good with nearly $84\%$ and hence close to the best result proposed 
in \cite{DBLP:conf/ismb/LinKB08}.}
\end{table}				

\begin{figure}
\centering
	\includegraphics[width=0.45\columnwidth,height=0.1\textheight]{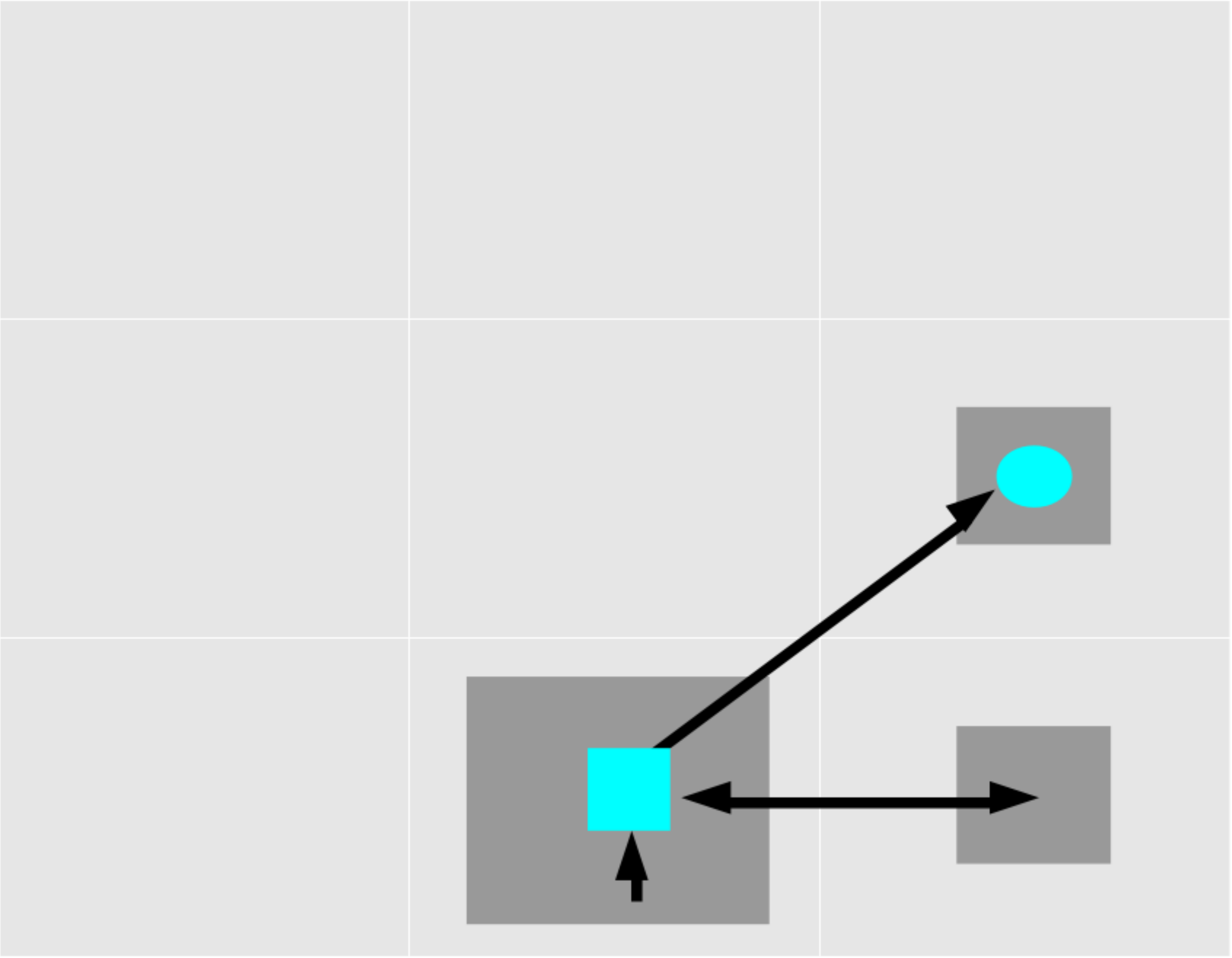}
	\includegraphics[width=0.45\columnwidth,height=0.1\textheight]{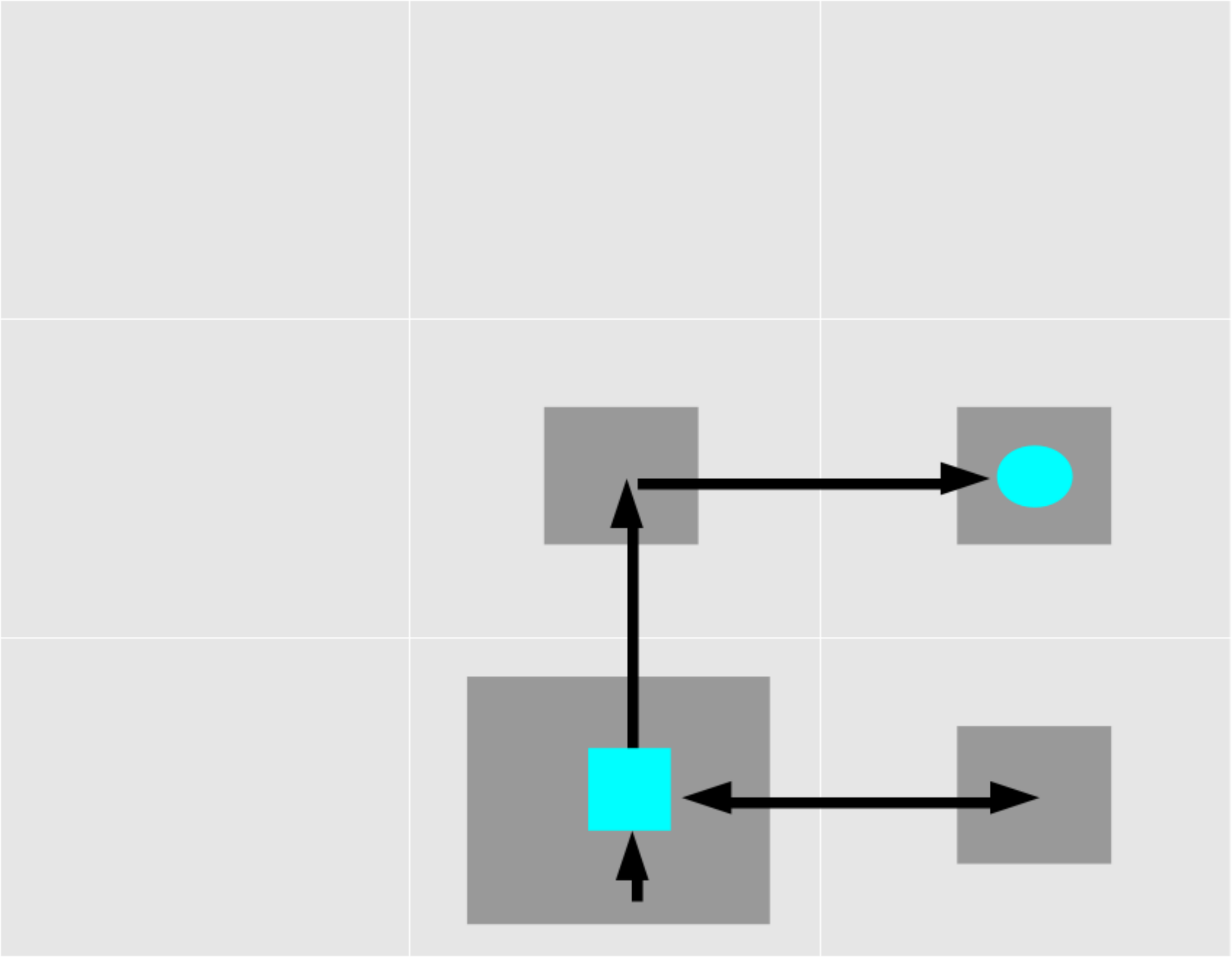}\\
	\includegraphics[width=0.9\columnwidth, height=0.1\textheight]{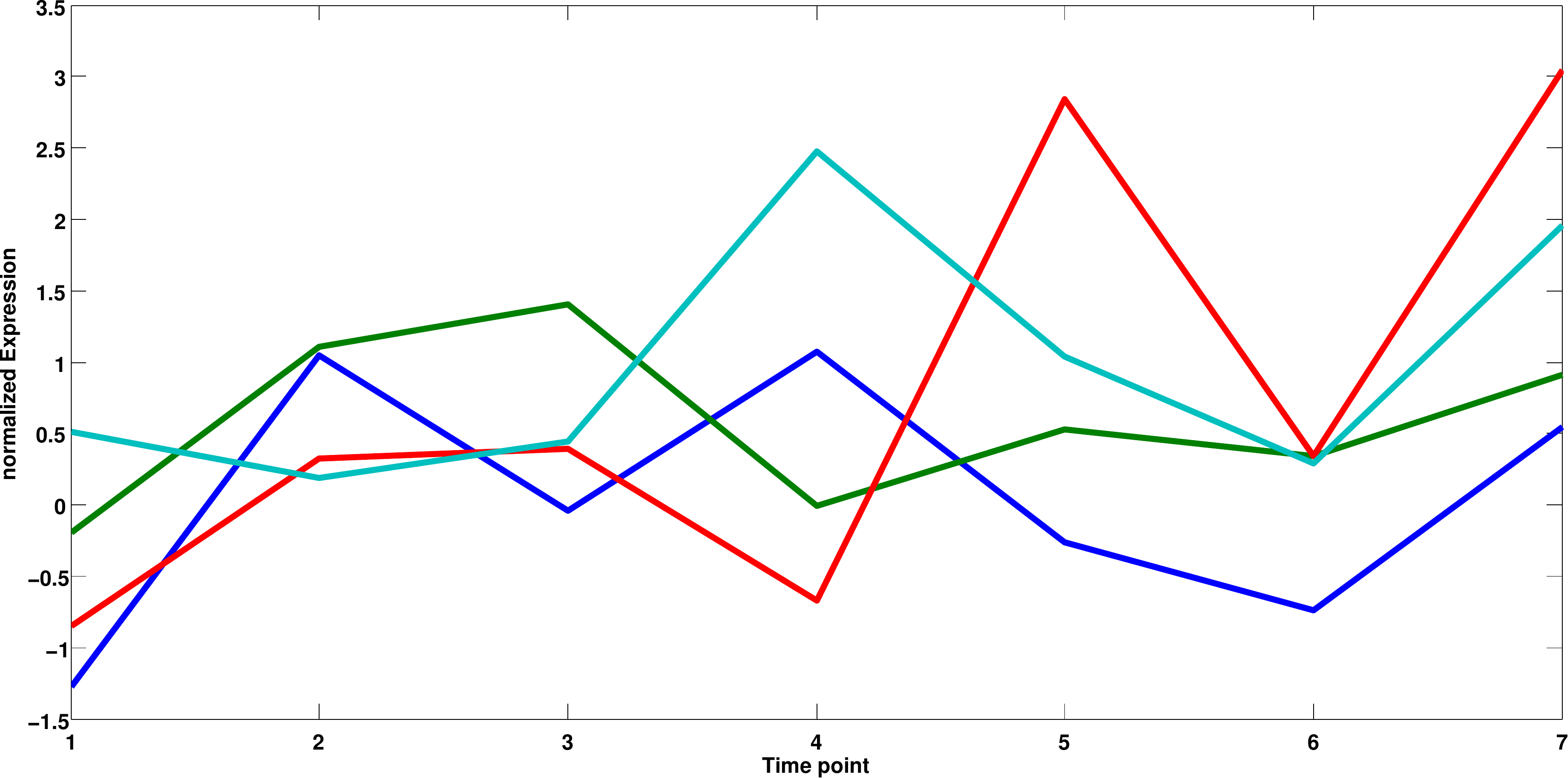}
	\caption{\label{fig:sgtmtt_map} Illustration of the $3 \times 3$ SGTM-TT mapping for the responder class. Plots in the first 
 	row show a typical state sequences. Also if the state sequences $Z$ are not identical we can expect that the underlying signals $X$
	are similar due to its close neighborhood on the map. This is reflected by such clustered signals at the bottom. The
	start of a sequence is indicated by $\square$ and the termination state by a $\circ$.}
\end{figure}

In Table \ref{tab:results_ms} we have summarized the prediction (test-set) results for the classification of the MS data set in comparison to the results
given in \cite{10.1371/journal.pbio.0030002}. The obtained mappings of the SGTM-TT are topology preserving\footnote{In our observations the topographic error
was reasonable small.} and we analyzed the mapping of the points to its prototypes and the neighborhoods. The map for the first class is depicted for two
temporal sequences in Figure \ref{fig:sgtmtt_map}. 

As expected, results improved by integration of feature selection or relevance learning compared to the full feature set.  
Overall the SGTM-TT with relevance learning performed very well and achieved good results of $92.5\%$ with respect to the best reported model
and also a smaller number of necessary features. 
\footnote{We would like to stress that due to the small sample size and the $4$ fold cross-validation a missclassification of $1$ point, accounts an error of $8\%$.}.   
Further the integrated relevance learning avoids multiple, time consuming runs within a wrapper approach like for the techniques used in 
\cite{DBLP:conf/ismb/LinKB08,DBLP:journals/bioinformatics/CostaSHS09}.
The obtained relevance profile is depicted in Figure \ref{fig:msrelevance} and provides direct access to an interpretation of the relevant features, or marker-candidates,
pruning irrelevant or noise dimensions. The values of the relevance profile are roughly gaussian distributed with $\mu=0.1$. We calculate a threshold $\zeta$ 
for the most relevant features using $\zeta=\mu+\sigma$ and obtain $7$ most relevant features, summarized in Table \ref{tab:results_ms_gene}.
\begin{table}[!t]
\begin{center}
\footnotesize
\begin{tabular*}{\columnwidth}{@{\extracolsep{\fill}}l|c|c|c}\hline
	Genes           & Relevance & found by Lin (7) & found by Costa (17) \\\hline\hline
 	MAP3K1		& 0.3014    &     X        &       X		\\           
    	NFkBIB		& 0.2609    &     -        &       - 		\\
    	IRF8		& 0.2584    &	  -        &       X		\\
    	Caspase 10	& 0.2471    &     X        &       X		\\
    	Jak2		& 0.1869    &     X        &       X		\\
    	FLIP		& 0.1842    &     -        &       -		\\
    	RIP		& 0.1647    &     -        &       -		
\end{tabular*}
\end{center}
\caption{\label{tab:results_ms_gene} Most relevant genes using SGTM-TT with relevance learning.}
\end{table}
                      
The SGTM-TT also inherently models different subgroups by the probabilistic regularizing model of the GTM and GTM-TT \cite{DBLP:journals/neco/BishopSW98,gtmttphd}. Hence the model complexity
is not so critical provided the map is reasonable large. This is a plus with respect to the approach presented in \cite{DBLP:journals/bioinformatics/CostaSHS09} which has the number
of groups as an additional meta parameter. 

\section{Conclusion}
We have presented a theoretically sound approach for the analysis of short temporal sequences. It is based on the novel idea to introduce supervision and relevance learning
into Generalized Topographic Mapping through time. Our results show that we are able to achieve improved or similar performance to alternative methods for the simulated 
and the MS data set. Further the prototype concept of the underlying method permits a better understanding of the model and extended visualization performance. We also obtain
a direct ranking of the individual features employing the relevance profile, rather by use of wrapper techniques. In future work we will explore more advance metric adaptation
schemes and alternative functional distance measures. Further we would like to apply our approach to non-clinical data and make it more flexible with respect to missing values.

\section*{Acknowledgment}
The authors thank: Peter Tino, University of Birmingham for interesting discussions about probabilistic
modeling and support during the early stage of this project and Falk Altheide, University of Bielefeld and
Tien-ho Lin, Carnegie Mellon University, USA for support with the simulation data. We would
also give extra thanks to Ivan Olier, University of Manchaster, UK; Iain Strachan, AEA Technology, Harwell, UK
and Markus Svensen, Microsoft Research, Cambridge, UK for providing code and support with the GTM and GTM-TT.

\paragraph{Funding:} {\footnotesize This work was supported by the German Res. Fund. (DFG),
HA2719/4-1 (Relevance Learning for Temporal Neural Maps) and by the Cluster of Excellence 277
Cognitive Interaction Technology funded in the framework of the German Excellence Initiative.}
\bibliographystyle{plain}
\bibliography{time,babs,references,pub_fms,pattern_neu}
\end{document}